\documentclass{isprs} % isprs class modified 23-04-2019 (Dennis Wittich)

\usepackage[numbers]{natbib}
\usepackage{subfigure}
\usepackage{setspace}
\usepackage{geometry} % added 27-02-2014 Markus Englich
\usepackage{epstopdf}
\usepackage[labelsep=period]{caption}  % added 14-04-2016 Markus Englich - Recommendation by Sebastian Brocks
\usepackage[british]{babel} 
\usepackage[hang]{footmisc}
\begin{document}

\title{Lightweight Road Environment Segmentation using Vector Quantization}
\date{}

% KAO: Remove extra spacing

% Anonymous submissions, authors' names should not be visible
% \author{
%  Orhan Altan\textsuperscript{1}, Ian Dowman\textsuperscript{2}, Florent Lafarge\textsuperscript{3}, Clément Mallet\textsuperscript{4}, Christian Heipke\textsuperscript{5} }
\author{Jiyong Kwag, Alper Yilmaz, Charles Toth}

% KAO: Remove extra newline
% Anonymous submissions, authors' affiliations should not be visible
%\address{
%	\textsuperscript{1 }ITU, Civil Engineering Faculty, 80626 Maslak Istanbul, Turkey - (oaltan, tozg, kulur, seker)@itu.edu.tr\\
%	\textsuperscript{2 }Dept.\ of Geomatic Engineering, University College London, Gower Street, London, WC1E 6BT UK - idowman@ge.ucl.ac.uk\\
%	\textsuperscript{3 }Université Côte d’Azur, INRIA – Sophia-Antipolis, France – florent.lafarge@inria.fr\\
%	\textsuperscript{4 }Univ. Gustave Eiffel, IGN-ENSG, LaSTIG – Saint-Mandé, France – clement.mallet@ign.fr\\
%	\textsuperscript{5 }Institute of Photogrammetry and GeoInformation, Leibniz Universit\"at Hannover, Germany - heipke@ipi.uni-hannover.de\\
%}
\address{Dept. of Civil Engineering, The Ohio State University, 281 W Lane Ave, Columbus, Ohio \\(kwag.3, yilmaz.15, toth.2)@osu.edu}

% If the corresponding author is NOT the final author, always add a % space before the subsequent comma, i.e.
% first author name\textsuperscript{a,}\thanks{Corresponding author} , % second author name \textsuperscript{b}, etc.
% thanks to Niclas Borlin 05-05-2016
% information on the corresponding author should not be used any longer and has been commented out
% C. Heipke, Jan 03,2024

% the use of the information of commissions and working groups should not be used any longer and has been commented out
% C. Heipke, Sept. 20,2022
%\commission{XX, }{YY} %This field is optional. If filled, XX and YY should be replaced by adequate numbers. See https://www2.isprs.org/commissions/
%\workinggroup{XX/YY} %This field is optional.
%\icwg{}   %This field is optional.

% KAO: Use times symbol
\abstract{ Road environment segmentation plays a significant role in autonomous driving. Numerous works based on Fully Convolutional Networks (FCNs) and Transformer architectures have been proposed to leverage local and global contextual learning for efficient and accurate semantic segmentation. In both architectures, the encoder often relies heavily on extracting continuous representations from the image, which limits the ability to represent meaningful discrete information. To address this limitation, we propose segmentation of the autonomous driving environment using vector quantization. Vector quantization offers three primary advantages for road environment segmentation. (1) Each continuous feature from the encoder is mapped to a discrete vector from the codebook, helping the model discover distinct features more easily than with complex continuous features. (2) Since a discrete feature acts as compressed versions of the encoder's continuous features, they also compress noise or outliers, enhancing the image segmentation task. (3) Vector quantization encourages the latent space to form coarse clusters of continuous features, forcing the model to group similar features, making the learned representations more structured for the decoding process. In this work, we combined vector quantization with the lightweight image segmentation model MobileUNETR and used it as a baseline model for comparison to demonstrate its efficiency. Through experiments, we achieved 77.0 \% mIoU on Cityscapes, outperforming the baseline by 2.9 \% without increasing the model's initial size or complexity. }


\maketitle

%\saythanks % added 28-02-2014 Markus Englich

\section{Introduction}\label{Introduction}

\sloppy

The High Definition (HD) map industry currently lacks a standardized structure or format, making it challenging to integrate different types of HD maps generated by various companies and researchers. However, segmentation-based maps are among the most widely used methods for generating precise HD maps. Segmentation offers several advantages. First, segmentation maps accurately separate drivable areas, sidewalks, and other essential road elements at a centimeter-level precision. Second, segmentation maps can be generated on-the-fly, independent of a central control system. Lastly, by incorporating existing offline bird’s-eye-view (BEV) maps, we can improve localization, detect anomalies in Position, Navigation, and Timing (PNT) systems, and mitigate these issues \cite{shi2019spatial}; \cite{isprsnavard}; \cite{zhu2021vigor}; \cite{zobar2024design}. This approach enhances the security of self-driving vehicles against GPS spoofing and hacking.

To achieve accurate segmentation of the road environment, Fully Convolutional Networks (FCNs) \cite{Long_2015_CVPR} were first proposed as a baseline model for semantic segmentation using a simple CNN-based architecture. FCNs extract low-level visual features through deep convolutional layers and perform segmentation with additional deep convolutional layers. However, CNN-based segmentation is limited by its use of only local features, lacking global semantic context throughout the image. To address this limitation, transformer-based architectures \cite{vaswani2017attention} have been proposed. Transformer models use attention mechanisms to introduce global context across the image, enabling powerful context learning \cite{dosovitskiy2020image}. Although the attention mechanism provides robust global contextual understanding, it lacks the fine spatial detail extraction that CNN-based architectures excel at. Thus, several models combine CNN and Transformer architectures to capitalize on each approach's strengths in segmentation tasks.
\begin{figure}[ht!]
\begin{center}
		\includegraphics[width=1.0 \columnwidth]{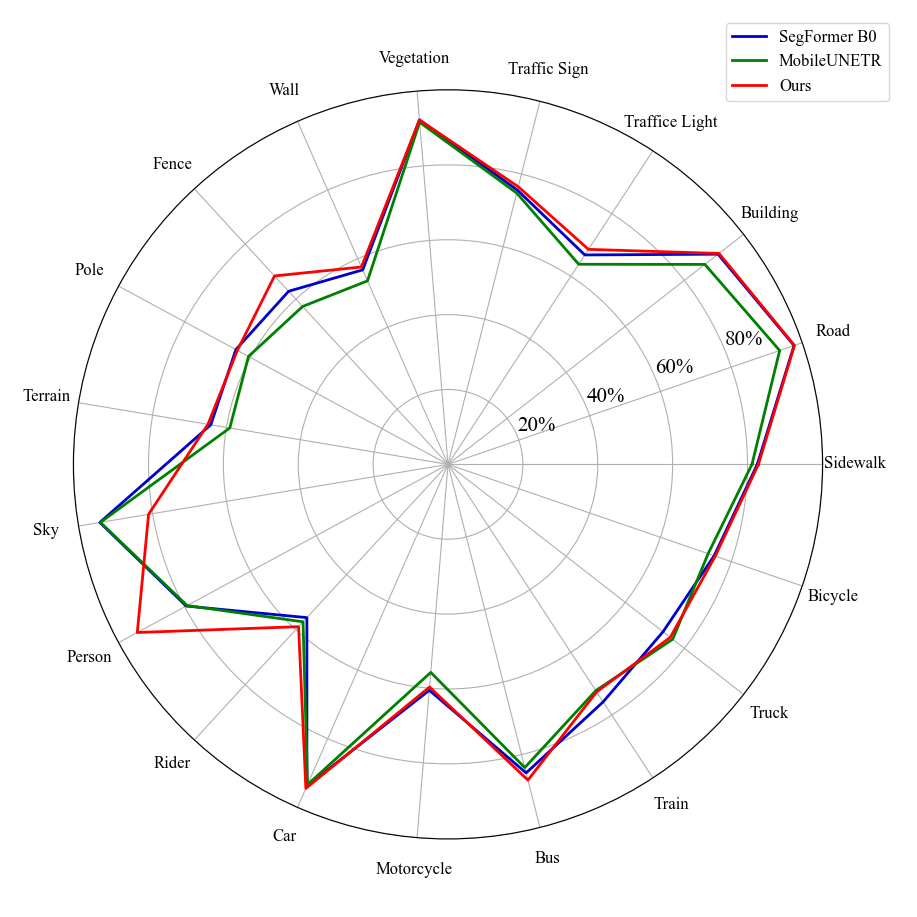}
	\caption{\textbf{Per-category segmentation IoU results on the Cityscapes validation set.} The graph displays the IoU evaluation for each category among SegFormer B0, MobileUNETR, and our proposed model. The results indicate that our architecture achieves better performance than the baseline models across 14 categories.}
\label{fig:figure2}
\end{center}
\end{figure} 

Despite the promising results achieved by combining CNN and Transformer architectures, the encoder is still focused primarily on extracting continuous features from the image. Vector Quantized Variational AutoEncoders (VQ-VAEs) \cite{van2017neural} first introduced discrete feature \begin{figure}[ht!]
\begin{center}
		\includegraphics[width=1.0 \columnwidth]{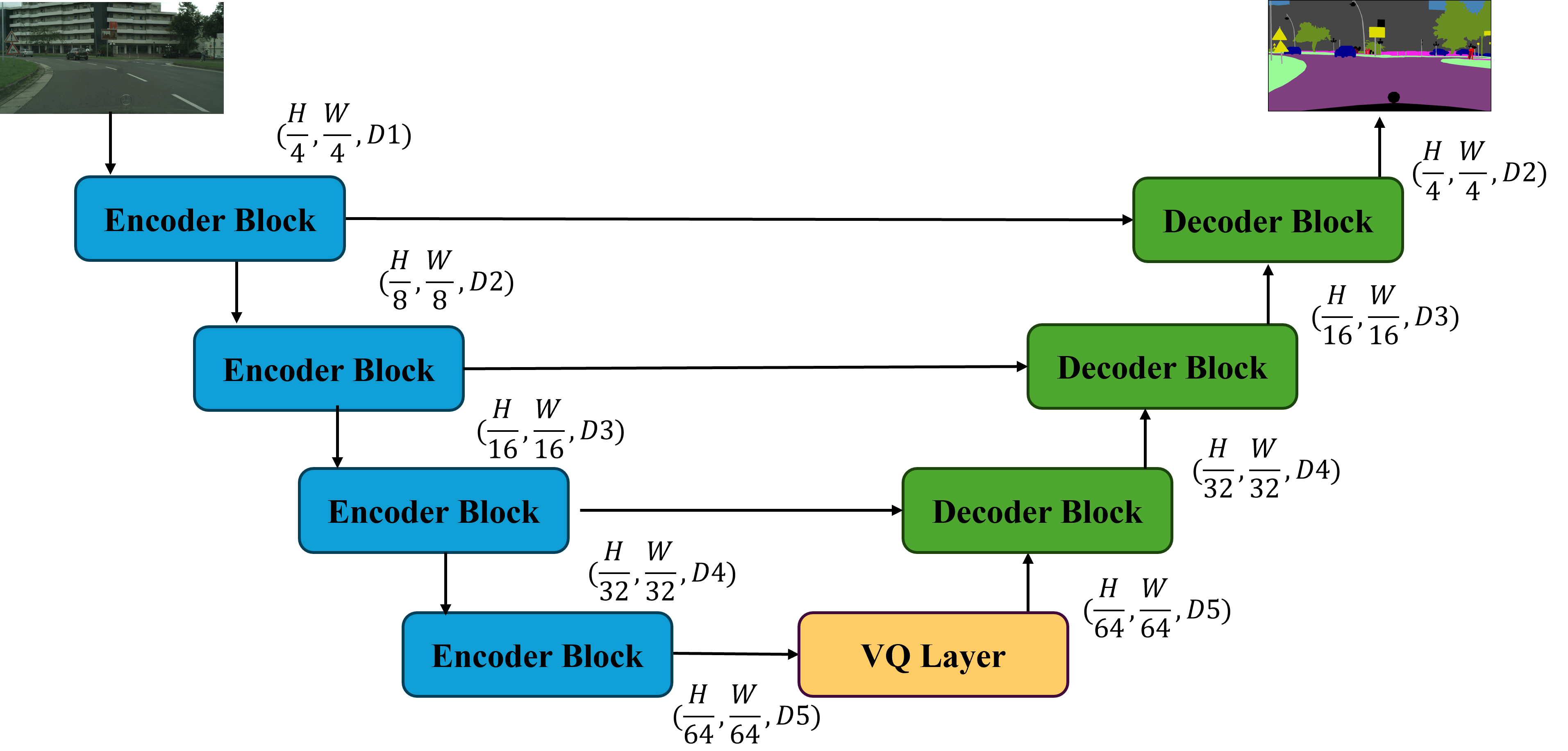}
	\caption{The proposed architecture combines MobileUNETR with vector quantization to transform the continuous representations extracted by the encoder into discrete representations. After the vector quantization layer, these discrete representations are processed by the MobileUNETR decoder to complete the segmentation task.}
\label{fig:figure3}
\end{center}
\end{figure}
extraction using vector quantization in generative models. In VQ-VAE, the primary goal is to learn a discrete feature representation of the continuous features in the latent space. This is achieved by mapping each continuous feature from the encoder to its closest predefined discrete feature. Through training, these discrete features learn to represent multiple continuous features. Discrete features naturally fit modalities such as language, speech, and images, performing better in complex reasoning, planning, and predictive learning tasks than continuous features, especially in generative models \cite{ho2020denoising}; \cite{rombach2022high}; \cite{navard2024knobgen}.

Inspired by the quantization of continuous features, we propose lightweight road environment segmentation using vector quantization. Although vector quantization has demonstrated robustness and effectiveness in the generative domain, its application in semantic segmentation remains underexplored \cite{santhirasekaram2022vector}; \cite{Gorade_2024_WACV}. By quantizing continuous features, vector quantization offers three primary advantages for segmentation tasks. First, the continuous features from the encoder are mapped to discrete vectors from the codebook, enabling the decoder to identify patterns more easily. Second, by using discrete features as compressed representations of continuous features, the model reduces noise, enhancing the segmentation task. Lastly, vector quantization encourages the latent space to form coarse clusters of continuous features, resulting in more structured representations for the decoder to process. In this work, we used MobileUNETR \cite{perera2024mobileunetr} as our baseline model for comparison. MobileUNETR is a lightweight image segmentation model that achieves state-of-the-art performance in medical image segmentation without requiring a large number of parameters. We simply added a vector quantization layer at the end of MobileUNETR encoder, aiming to map continuous features from the encoder to discrete features. Through our experiments, we achieved 77.0 \% mIoU on the Cityscapes dataset, showing a 2.9 \% improvement over the baseline while maintaining the model's original size and complexity.

\section{Related Work}\label{sec:Related Work}

\textbf{CNN-Based Segmentation.} Semantic segmentation, a crucial task in computer vision, has advanced significantly with the rise of deep learning techniques. Early approaches relied primarily on CNN, which has proven effective in capturing hierarchical features from images. FCNs \cite{Long_2015_CVPR}
\begin{figure}[ht!]
\begin{center}
		\includegraphics[width=1.0 \columnwidth]{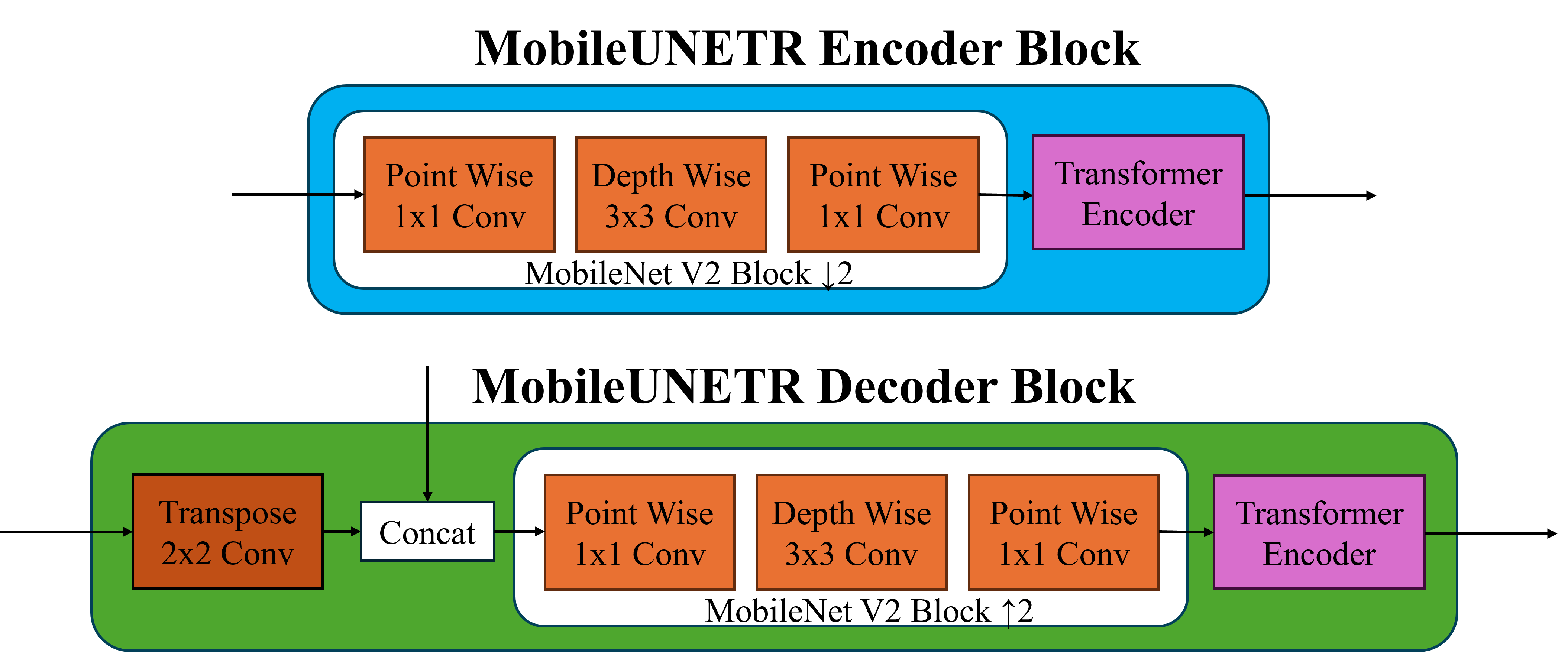}
	\caption{(Top) The MobileUNETR encoder utilizes a pretrained MobileViT encoder, composed of MobileNet V2 blocks and Transformer encoder blocks, for efficient feature extraction. (Bottom) The MobileUNETR decoder mirrors the encoder structure, but instead upsamples the features from the vector quantization layer for semantic segmentation.}
\label{fig:figure4}
\end{center}
\end{figure}
were among the first to adapt CNNs for pixel-wise labeling, replacing fully connected layers with convolutional layers to produce segmentation maps with the same dimensions as the input images. Subsequent CNN-based models like U-Net \cite{ronneberger2015u}, SegNet \cite{badrinarayanan2017segnet}, and DeepLab \cite{chen2017deeplab} introduced innovations such as encoder-decoder architectures, skip connections, and atrous convolutions to improve segmentation accuracy. These models have set benchmarks for various datasets, demonstrating the robustness of CNNs for semantic segmentation tasks.

\textbf{Transformer-Based Segmentation.} Recently, Vision Transformers (ViTs) \cite{dosovitskiy2020image} have become strong alternatives to CNNs. Transformers \cite{vaswani2017attention}, initially designed for natural language processing, have been used for image tasks by treating image patches as tokens. Models like Segmenter \cite{strudel2021segmenter}, SegFormer \cite{xie2021segformer}, and SegFormer3D \cite{Perera_2024_CVPR} leverage hierarchical Transformer encoders and lightweight MLP decoders to capture global context. Furthermore, models such as TransUNet \cite{chen2021transunet}, TransBTS \cite{wenxuan2021transbts}, and TransFuse \cite{zhang2021transfuse} combine CNN and Transformer architectures to improve the quality of semantic segmentation. These Transformer-based models excel at modeling long-range dependencies and have shown competitive results compared to CNN-based architectures.

\textbf{Vector Quantization.} Since the development of VQ-VAE as a foundational model for discrete representation learning, numerous works have focused on improving VQ-VAE from various perspectives. Residual VQ \cite{zeghidour2021soundstream} proposes using multiple discrete codebooks to recursively quantize the residuals of continuous representations after codebook matching. Group-residual vector quantization \cite{yang2023hifi} enables grouping in the feature dimensions, achieving equivalent results while using fewer codebooks. Improved VQGAN \cite{yu2021vector} introduces L2 normalization for codebook vectors and output vectors from encoder, effectively employing cosine similarity like distance measurements using MSE loss. More recently, Finite Scalar Quantization \cite{mentzer2023finite} aims to simplify vector quantization for generative modeling by eliminating commitment losses and codebook updates for discrete representation learning. This method rounds each scalar to discrete levels with straight-through gradients, showing improved performance over VQ-VAE, especially when the number of codebooks is significantly increased.
\begin{figure}[ht!]
\begin{center}
		\includegraphics[width=0.7 \columnwidth]{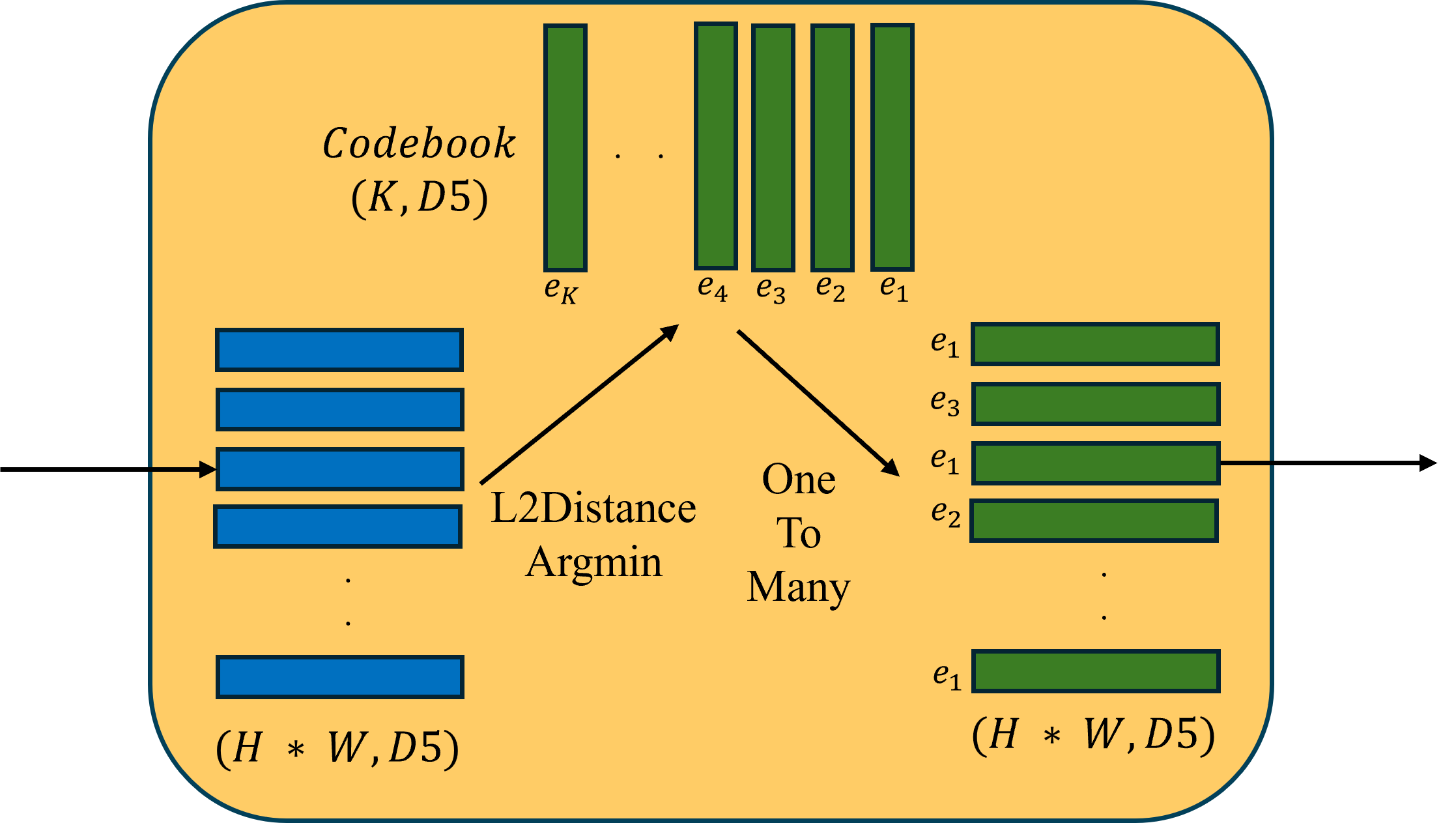}
	\caption{Simplified representation of the vector quantization layer: each continuous representation from the encoder is mapped to the nearest predefined discrete feature from the codebook, which is then used as input to the decoder.}
\label{fig:figure5}
\end{center}
\end{figure}
\section{Methodology}\label{sec:Methodology}

\subsection{MobileUNETR Encoder}\label{sec:MobileUNETR_Encoder} MobileUNETR is an encoder-decoder-based architecture that leverages both CNN and Transformer components. It follows the same overall structure as U-NET, where each encoder block is connected to its corresponding decoder block via skip connections. Figure 1 provides a high-level representation of MobileUNETR. Using both CNN and Transformer architectures, the main goal is to simultaneously incorporate local and global feature extraction for the segmentation task. Furthermore, by adopting the MobileNet V2 \cite{sandler2018mobilenetv2} block with MobileViT \cite{mehta2021mobilevit} encoder ideas in on both the encoder and the decoder, MobileUNETR achieves a lightweight architecture with state-of-the-art accuracy. Specifically, MobileUNETR uses an ImageNet-1K pre-trained MobileViT as its encoder.

As shown in Figure 2, the MobileUNETR encoder consists of two components: the MobileNet V2 block and the Transformer encoder block. The main purpose of the MobileNet V2 block is to downsample and extract local features from the image. It employs two types of convolutions: 1x1 pointwise convolution and 3x3 depthwise convolution. Point-wise convolution projects the tensor to a higher dimension, interacting only within its dimension, while depth-wise convolution encodes local spatial features around individual pixels through a 3x3 kernel. After the MobileNet V2 block extracts the features, the Transformer encoder performs self-attention followed by a feedforward network. However, the self-attention mechanism of the Transformer encoder differs from that of a vanilla Transformer. The input tensor \(X\in{R}^{H\times W\times\ d}\) is split into \(N\) flattened patches \(X_p\in{R}^{P\times N\times\ d}\), where \(P=wh\)  is the patch size, and \(N=\frac{HW}{P}\) is the number of patches. Self-attention is then performed to capture the interpatch relationships, after which \(X_p\) is folded back to the original shape \(X\in{R}^{H\times W\times\ d}\). Using interpatch self-attention, the Transformer encoder achieves a lightweight architecture that reduces computational cost while capturing global context.    

\subsection{Vector Quantization}\label{sec:Vector_Quantization}

Following the MobileUNETR encoder, a vector quantization layer maps the output features \(X\in{R}^{H\times W\times\ d}\) from the encoder to discrete features. First, we define a discrete latent embedding called a codebook \(e\in{R}^{K\times d}\) where \(K\) represents the number of vectors in the codebook. The output feature \(X\) is then quantized based on its distance to each vector in the codebook \(e_k, k \in 1...K\). As a result, each vector \(X\) is replaced by the nearest index of the codebook vector.
\begin{equation}\label{equ:1}
	Quantize(X) = e_k \quad where \quad k = \arg\min_{j} \|X - \mathbf{e}_j\|,
\end{equation}
However, by matching the nearest index of the codebook vectors in Equation (1), no gradient updates occur in the codebook or encoder. Therefore, the straight-through estimator is used to copy the gradient from the decoder back to the encoder.

After mapping discrete vectors from the codebook to the encoder output features, it is necessary to learn a discrete representation from the continuous encoder output. In VQ-VAE, two types of vector quantization losses are introduced: codebook loss and commitment loss. Codebook loss applies only to the codebook, updating the vectors in the codebook to be closer to the encoder’s continuous features. The loss of commitment applies only to the encoder, encouraging the continuous features of the encoder to remain close to the chosen codebook vector. This ensures that the mapped continuous vectors and discrete vectors from the codebook do not fluctuate excessively between discrete vectors during training.
\begin{equation}\label{equ:2}
	\mathcal{L_{VQ}} = \|sg[X] - \mathbf{e}\|_2^2 + \beta \|sg[\mathbf{e}] - X\|_2^2,
\end{equation}
where sg represents the stop-gradient, which acts as an identity function during the forward pass but has zero partial derivatives. Thus, the operand is treated as a constant that is not updated. The first term in \(\mathcal{L_{VQ}}\) represents the codebook loss, while the second term represents the commitment loss. The weight \(\beta\) in the commitment loss limits the encoder’s focus on updating the similarity between the output features and the codebook vectors.

\subsection{MobileUNETR Decoder}\label{sec:MobileUNETR_Decoder} 
After mapping the continuous features of the encoder to discrete vectors in the codebook, MobileUNETR maintains a consistent design with its encoder architecture. Previous models typically used deep convolutional networks in their decoders to perform segmentation tasks. While effective, these are computationally expensive and require a heavyweight decoder architecture. To address this, MobileUNETR uses the MobileNet V2 block for upsampling, along with a Transformer encoder that leverages interpatch self-attention for semantic segmentation.

As illustrated in Figure 3, the decoder process begins by upsampling the feature map from the previous layer using transpose convolution, which is then concatenated with the corresponding upsampled feature map from the encoder via skip connections. This design allows MobileUNETR to progressively increase spatial resolution while refining features at each stage using a hierarchical structure. The MobileNet V2 block refines local features through a combination of point- and depth-wise convolutions, enhancing computational efficiency. Finally, the Transformer encoder employs interpatch self-attention to globally refine these features, allowing MobileUNETR to achieve accurate segmentation results without relying on computationally heavy deep convolutional decoders. This results in a lightweight, efficient architecture suitable for real-time applications.

For the semantic segmentation loss function, we used the commonly applied cross-entropy loss. The cross-entropy loss calculates the loss for each pixel by comparing the predicted class
\begin{figure*}[ht!]
\begin{center}
		\includegraphics[width=2.1\columnwidth]{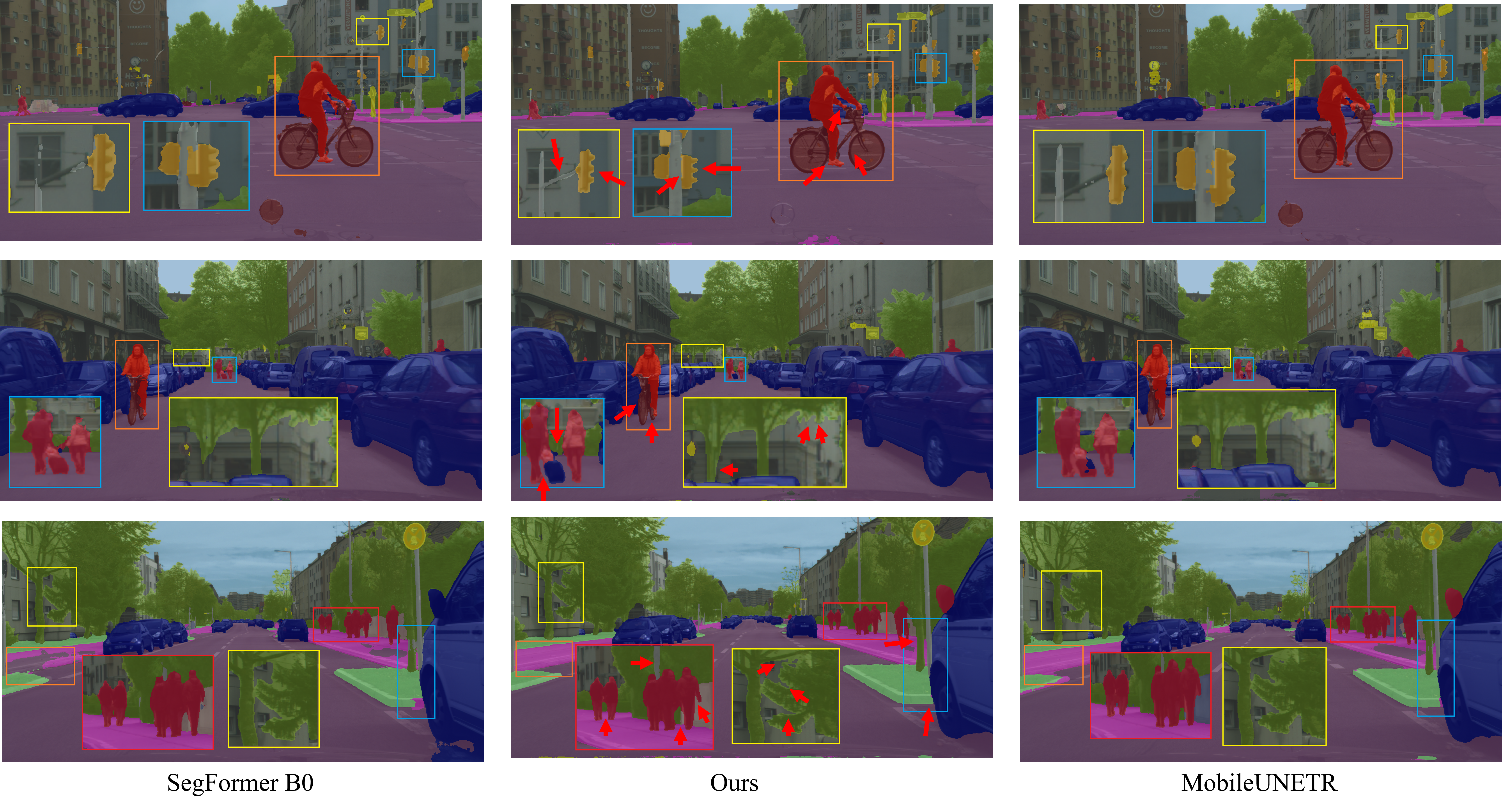}
	\caption{\textbf{Visualization results on Cityscapes.} Compared to the baseline model MobileUNETR (Right), our model (Center) predicts segmentation with more precise object edges. We also provide a comparison with SegFormer B0 (Left), showing that, despite a smaller size and lower FLOPs, vector quantization enhances performance over the baseline and outperforms SegFormer B0.}
\label{fig:figure1}
\end{center}
\end{figure*}
probabilities with the ground truth class. Our final loss function for the model is as follows:
\begin{equation}\label{equ:3}
	\mathcal{L} = \mathcal{L_{CE}} + \mathcal{L_{VQ}},
\end{equation}

\section{Experiments}\label{sec:Experiments}

\subsection{Dataset}\label{sec:Dataset} We use the publicly available Cityscapes road environment dataset. Cityscapes contains various driving scenes for semantic segmentation, with training, validation, and test sets consisting of 3,000, 500, and 1,500 high-resolution images, respectively, each with corresponding ground truth fine segmentation annotations. The Cityscapes dataset was collected across 50 cities in Germany under various conditions, including different months, weather variations, and day and night scenes. Cityscapes includes 19 categories representing the road environment.

\subsection{Implementation Details}\label{sec:Implementation_Details}

Our architecture was implemented using the mmsegmentation codebase in PyTorch and trained on a server with three RTX A6000 GPUs. We used a pre-trained MobileViT encoder on the ImageNet-1K dataset and initialized the decoder randomly. For the vector quantization layer, the codebook embeddings were uniformly initialized within the range \([-1/K, 1/K]\), where \(K\) is the number of vectors in the codebook. This setup ensures stable training by preventing the codebook vectors from becoming overly dispersed or fluctuating excessively across dimensions. 0.25 is selected as \(\beta\) for the commitment loss. For a fair comparison, we followed the same training pipeline as SegFormer and MobileUNETR. Data augmentation was applied with a random resize ratio between 0.5 and 2.0, random cropping to 1024x1024, and horizontal flipping. Our model was trained using the AdamW optimizer for 160K iterations with a batch size of 9.

The initial learning rate was set to 0.000006, followed by a polynomial learning rate scheduler with a factor of 1.0. As discussed in the methodology, we used cross-entropy loss for segmentation along with vector quantization loss as an auxiliary loss to update both the codebook and the encoder. The number of codebook vectors used was 19, chosen for reasons detailed in Section 4.4. During validation, we used a sliding window test with 1024x1024 windows. To evaluate segmentation performance, we provided both visualization results, mean Intersection over Union (mIoU), and class-by-class Intersection over Union (IoU) compared with the baseline models.

\subsection{Results}\label{sec:Results}

We conducted experiments on the Cityscapes dataset, using MobileUNETR and SegFormer B0 as our baseline models for comparison. MobileUNETR is a state-of-the-art medical segmentation model known for its lightweight architecture and accuracy comparable to much larger models. SegFormer B0 is a foundational model for road environment segmentation, featuring a lightweight design with a Transformer encoder and a simple MLP decoder.

Figure 5 provides a visualization comparison between our proposed model and the baseline models, focusing on key segments of the results. The close-up comparisons show that our model achieves more refined boundaries, particularly in segments representing persons, vegetation, roads, and sidewalks. Table 1 presents comparisons with the baseline models. Our model improved MobileUNETR’s mIoU by 2.9 \% while 
\begin{table*}[h]
	\centering
        \begin{tabular}{|l|c|c|c|c|c|c|c|}
          \hline
          Method & Iterations & Loss Function & Encoder ↓ & Decoder ↓ & Total ↓ & FLOPs ↓ & mIoU ↑ \\
           &  &   & (\# params) & (\# params) & (\# params) &  &  \\
          \hline
          SegFormer B0 & 160K & Cross Entropy & 3.4 M & 0.4 M & 3.8 M & 17.7 G & 76.2 \% \\
          MobileUNETR & 320K & Cross Entropy + Dice & 1.3 M & 1.7 M & 3.0 M & 1.3 G & 75.9 \% \\
          MobileUNETR & 160K & Cross Entropy & 1.3 M & 1.7 M & 3.0 M & 1.3 G & 74.1 \% \\
          \hline
          \textbf{Ours} & 160K & Cross Entropy & 1.3 M & 1.7 M & 3.0 M & 1.3 G & \textbf{77.0 \%} \\
          \hline
        \end{tabular}
	\caption{\textbf{Comparison to baseline methods on Cityscapes validation set.} Our model demonstrates significant advantages in terms of total parameters, FLOPs, and accuracy, using a 1024 x 1024 resolution across all three models.}
\label{tab:Margin_settings3}
\end{table*}
maintaining the original size and FLOPs of the architecture. MobileUNETR achieved comparable mIoU only after incorporating Dice loss and doubling the iterations to 320K. Furthermore, our work exceeds SegFormer B0 by 0.8 \% even with a lower number of parameters and FLOPs. Figure 1 and Table 2 provide detailed graphs and tables of the results per category compared to the baselines. Our model exceeds the baselines in 14 categories while retaining high accuracy across others. In particular, our model outperformed both baselines in person segmentation by 15 \%, an essential feature for driving safety in autonomous vehicles compared to categories such as sky and terrain.
\begin{table}[h]
	\centering
		\begin{tabular}{|l|c|c|c|}\hline
			 Category&SegFormer B0&MobielUNETR&\textbf{Ours}\\\hline
			 Road&97.73&93.67&\textbf{97.85}\\
			 Sidewalk&82.62&81.2&\textbf{82.92}\\
              Building&91.41&86.96&\textbf{91.73}\\
              Wall&56.73&53.53&\textbf{57.57}\\
              Fence&62.85&57.33&\textbf{68.42}\\
              Pole&\textbf{64.43}&60.59&63.98\\
              Traffic Light&66.82&63.85&\textbf{68.6}\\
              Traffic Sign&75.56&74.76&\textbf{76.49}\\
              Vegetation&\textbf{92.32}&91.7&\textbf{92.32}\\
              Terrain&64.24&59.09&\textbf{65.01}\\
              Sky&\textbf{94.29}&94.14&81.09\\
              Person&79.51&79.24&\textbf{94.37}\\
              Rider&55.65&57.17&\textbf{58.93}\\
              Car&93.84&93.43&\textbf{94.52}\\
              Truck&72.79&\textbf{76.06}&75.24\\
              Bus&85.02&83.56&\textbf{87.0}\\
              Train&\textbf{75.78}&72.19&72.6\\
              Motorcycle&\textbf{60.55}&55.76&59.69\\
			 Bicycle&75.17&73.47&\textbf{75.48}\\\hline
		\end{tabular}
	\caption{\textbf{Per-category segmentation IoU results on the Cityscapes validation set.} The table presents the IoU evaluation for each category among SegFormer B0, MobileUNETR, and our model. Bold values indicate the highest IoU in each category. The results show that our proposed architecture outperforms the baseline models in 14 categories.}
\label{tab:Margin_settings1}
\end{table}
\subsection{Ablation Study}\label{sec:Ablation_Study}

\textbf{Codebook Usage.} In this ablation study, we experimented with different numbers of vectors in the codebook. In VQ-VAE, increasing the size of the codebook by a factor of two results in an exponential decrease in FID until the codebook size reaches \(2^{10}\), after which the improvements plateau. We followed a similar approach to analyze the effect of codebook size on segmentation performance. Furthermore, a significant concern with increasing the codebook size is the potential impact on codebook usage. Although increasing the size of the codebook can improve accuracy, it can also reduce the use of the codebook, leading to unused parameters. Since one of our goals is to maintain a lightweight architecture, we also examined the codebook perplexity as the number of vectors increased. 

We experimented with three different codebook sizes: \(19\), \(19 \times 5\), \(19 \times 10\). These sizes were chosen because we are focusing on predicting 19 distinct classes, and we wanted to explore the effect of varying the codebook's capacity relative to the number of classes. As shown in Table 3, the mIoU remains relatively stable between these different codebook sizes, with only minor variations of around 0.5\%. This small fluctuation in mIoU likely falls within the expected range due to random model initialization rather than an effect of the codebook size itself.

Furthermore, the codebook utilization rate remains consistently 100\% in all sizes tested, indicating that every vector within the codebook is effectively used during training. This full usage suggests that the model distributes representations effectively within the available codebook entries, regardless of the size. Therefore, increasing the codebook size does not necessarily improve performance, but still ensures that all codebook entries contribute to the model's predictions. This observation underscores that a compact codebook can be sufficient for our class prediction task while maintaining full utilization, thereby reinforcing the efficiency of using a codebook closely aligned with the number of target classes.
\begin{table}[h]
	\centering
		\begin{tabular}{|l|c|c|}\hline
			 Codebook Size& mIoU ↑& USG ↑\\\hline
			 19  &\textbf{ 77.0\%} & 100\% \\
              \(19 \times 5\)&  76.5\%&  100\% \\
              \(19 \times 10\)&  76.6\% & 100\% \\\hline
		\end{tabular}
	\caption{\textbf{Ablation Study on Codebook Size.} The table presents mIoU and codebook utilization for various codebook sizes. The mIoU does not change significantly across these sizes, and the usage rate remains at 100 \%. }
\label{tab:Margin_settings2}
\end{table}

\section{Conclusion}\label{sec:Conclusion}

In this work, we presented a novel approach to lightweight road environment segmentation by integrating vector quantization with MobileUNETR. By mapping continuous features to discrete codebook representations, our method enhances segmentation precision while preserving model efficiency. Experiments on the Cityscapes dataset demonstrated that our model achieves a 2.9 \% improvement in mIoU over baseline model, with especially notable gains in person segmentation, a critical feature for autonomous driving safety. Our approach maintains the original model size and complexity, making it well-suited for real-time applications. Future work could explore further optimizations in codebook design on skip connection. Thus, we can anticipate that incorporating hierarchical vector quantization layers within the decoder blocks will enhance performance, while still meeting the demands for lightweight, high-accuracy segmentation in autonomous driving.

\section*{Acknowledgments}

This work was supported in part by the U.S. Department of Transportation under Grant 69A3552348327 for the CARMEN+ University Transportation Center.

\bibliography{reference} % Include your own bibliography (*.bib), style is 

\end{document}